%% file: main.tex
\pdfoutput=1
\documentclass{article}

\usepackage{microtype}
\usepackage{graphicx}
\usepackage{subcaption}
\usepackage{booktabs} 

\usepackage{hyperref}

\usepackage[preprint]{icml2026}

\makeatletter
\renewcommand{\ICML@appearing}{}

\makeatother

\usepackage{amsmath}
\usepackage{amssymb}
\usepackage{mathtools}
\usepackage{amsthm}
\usepackage{enumitem}
\usepackage[capitalize,noabbrev]{cleveref}

\theoremstyle{plain}

\theoremstyle{definition}

\theoremstyle{remark}

\icmltitlerunning{JOintGS: Joint Optimization for In-the-Wild Monocular Reconstruction}

\usepackage{xspace}
\makeatletter
\DeclareRobustCommand\onedot{\futurelet\@let@token\@onedot}
\def\@onedot{\ifx\@let@token.\else.\null\fi\xspace}
\def\eg{\emph{e.g}\onedot}

\makeatother

\usepackage[table]{xcolor}

\begin{document}

\twocolumn[
  \icmltitle{JOintGS: Joint Optimization of Cameras, Bodies and 3D Gaussians for In-the-Wild Monocular Reconstruction}

  \icmlsetsymbol{equal}{*}
  \icmlsetsymbol{cor}{$\dagger$}

  \begin{icmlauthorlist}
    \icmlauthor{Zihan Lou}{equal,whu}
    \icmlauthor{Jinlong Fan}{equal,hdu}
    \icmlauthor{Sihan Ma}{ntu}
    \icmlauthor{Yuxiang Yang}{hdu}
    \icmlauthor{Jing Zhang}{whu,cor}
  \end{icmlauthorlist}

  \icmlaffiliation{whu}{School of Computer Science, Wuhan University, China}
  \icmlaffiliation{hdu}{Hangzhou Dianzi University, China}
  \icmlaffiliation{ntu}{Nanyang Technological University, Singapore}

  \icmlcorrespondingauthor{Jing Zhang}{jingzhang.cv@gmail.com}

  \icmlkeywords{3D Human Reconstruction, 3D Gaussian Splatting, Monocular Reconstruction, Computer Vision}

  \vskip 0.3in
]

\printAffiliationsAndNotice{\icmlEqualContribution $\dagger$ Corresponding author. }

\input{sec/0_abstract}
\input{sec/1_intro}

\input{sec/2_relatedworks}

\input{sec/3_method}

\input{sec/4_Experiments}

\input{sec/5_conclusion}

\bibliography{main}
\bibliographystyle{icml2026}

\newpage
\appendix
\onecolumn

\renewcommand{\thefigure}{S\arabic{figure}}
\setcounter{figure}{0}
\renewcommand{\thetable}{S\arabic{table}}
\setcounter{table}{0}

\input{sec/X_suppl}

\end{document}

%% file: sec/0_abstract.tex
\begin{abstract}
Reconstructing high-fidelity animatable 3D human avatars from monocular   
RGB videos remains challenging, particularly in unconstrained in-the-wild   
scenarios where camera parameters and human poses from off-the-shelf methods   
(e.g., COLMAP, HMR2.0) are often inaccurate.
Splatting (3DGS) advances demonstrate impressive rendering quality and   
real-time performance, they critically depend on precise camera calibration   
and pose annotations, limiting their applicability in real-world settings.
We present \textbf{JOintGS}, a unified framework that jointly optimizes   
camera extrinsics, human poses, and 3D Gaussian representations from   
coarse initialization through a \textit{synergistic refinement mechanism}.  
Our key insight is that explicit foreground-background disentanglement   
enables mutual reinforcement: static background Gaussians anchor camera   
estimation via multi-view consistency; refined cameras improve human body   
alignment through accurate temporal correspondence; optimized human poses   
enhance scene reconstruction by removing dynamic artifacts from static   
constraints.
We further introduce a temporal dynamics module to capture   
fine-grained pose-dependent deformations and a residual color field to   
model illumination variations.
Extensive experiments on NeuMan and EMDB   
datasets demonstrate that JOintGS achieves superior reconstruction quality,   
with 2.1~dB PSNR improvement over state-of-the-art methods on NeuMan dataset, while maintaining real-time rendering.   
Notably, our method shows significantly enhanced robustness to noisy initialization compared to the baseline.
Our source code is available at \url{https://github.com/MiliLab/JOintGS}.
\end{abstract}

%% file: sec/1_intro.tex
\input{figures/Overview_Framework}
\section{Introduction}
\label{sec:Introduction}
Reconstructing high-fidelity, animatable 3D human avatars from monocular videos 
has emerged as a fundamental challenge in computer vision with broad applications 
in virtual reality, telepresence, digital entertainment, and human-computer 
interaction~\cite{wang2024survey}. Recent advances in neural rendering, particularly 
3D Gaussian Splatting (3DGS)~\cite{kerbl20233d}, have demonstrated unprecedented 
rendering quality and efficiency, enabling real-time photorealistic synthesis. 
Building upon this success, several methods~\cite{qian20243dgs,moon2024expressive,guo2025vid2avatar,hu2024gauhuman,guo2023vid2avatar,kocabas2024hugs,zhang2025odhsr, li2024gaussianbody, shao2024splattingavatar} have 
extended 3DGS to dynamic human reconstruction, achieving impressive results 
on controlled datasets with multi-view captures or precisely calibrated cameras.

However, these methods face a fundamental   
limitation when applied to \textit{in-the-wild} monocular videos: they   
critically depend on highly accurate camera parameters and human pose   
annotations to maintain consistent spatial-temporal alignment. This dependency   
severely limits practical applicability, as obtaining such precise estimates   
remains notoriously challenging in unconstrained settings. Traditional   
Structure-from-Motion (SfM) pipelines like COLMAP~\cite{schoenberger2016sfm,schoenberger2016mvs}  
often struggle with dynamic scenes, producing noisy camera trajectories due   
to insufficient static feature correspondences. Similarly, monocular human   
pose estimators~\cite{shan2022p, xu2023auxiliary, zhang2023pymaf,cai2023smpler, goel2023humans}, while achieving impressive 2D keypoint   
detection, frequently yield inaccurate 3D SMPL parameters due to depth   
ambiguity and occlusions. Even modest errors in these prerequisites cascade into severe artifacts such as inaccurate alignment, temporal inconsistencies, and unrealistic human-scene interpenetration. 

The key insight is that while obtaining precise camera and pose parameters is difficult, coarse estimates are readily derived. Rather   
than treating these initial estimates as fixed ground truth, we pose a central question:   
\textit{Can we leverage the rich geometric and photometric constraints   
inherent in 3DGS-based reconstruction to jointly refine camera   
trajectories, human poses, and 3D representations?} This question motivates 
our proposed framework, \textbf{JOintGS}, which formulates dynamic human reconstruction as a unified optimization problem.

JOintGS introduces a \textit{synergistic refinement mechanism} through explicit foreground-background disentanglement,  where different components mutually reinforce each other through three complementary pathways: (1) Static background Gaussians, remaining consistent across frames, naturally provide multi-view geometric constraints for camera pose estimation. By exploiting photometric consistency on static regions, we progressively refine   
camera trajectories without being affected by dynamic human motion; (2) With refined cameras establishing accurate spatial-temporal correspondences, we optimize human poses to minimize re-projection errors of human silhouettes and appearance, correcting initialization errors from monocular pose estimators; (3) Improved camera and poses enhance human-scene disentanglement by providing accurate   
foreground-background separation. Clean background constraints, in turn, stabilize camera estimation by removing dynamic artifacts that violate static scene assumptions. 
Unlike previous methods that either treat camera and pose as fixed inputs or optimize them separately~\cite{jiang2023instantavatar,hu2024gaussianavatar} (Figure~\ref{fig:diff_others}), our approach forms a closed-loop system where each component progressively corrects errors in the others. This synergistic design enables robust reconstruction from coarse initialization without requiring pre-calibrated inputs or expensive preprocessing.  

Furthermore, to effectively model the complex dynamics of human motion, we introduce two complementary components: a temporal offset module that learns per-frame non-rigid geometric deformations to capture fine-grained changes like clothing wrinkles, and a residual color field that models appearance variations caused by lighting changes and view-dependent effects. These modules enable our method to faithfully reconstruct both geometric and photometric details that are challenging to capture with canonical representations alone~\cite{weng2022humannerf,hu2024gaussianavatar,qian20243dgs}.

We perform comprehensive evaluations on two challenging in-the-wild datasets, NeuMan~\cite{jiang2022neuman} and EMDB~\cite{kaufmann2023emdb}. Experimental results show that JOintGS delivers superior reconstruction quality, achieving a 2.2 dB PSNR improvement over SOTA approaches on the NeuMan dataset, while maintaining real-time rendering performance. Furthermore, JOintGS demonstrates stronger robustness to noisy initialization, exhibiting only a 0.9 dB PSNR drop at $\sigma{=}0.01$, in contrast to the 3.7 dB drop observed with the strong baseline HUGS~\cite{kocabas2024hugs}. Comprehensive ablation studies further confirm the necessity of the joint optimization strategy and the effectiveness of each component in our synergistic refinement mechanism.

\noindent In summary, our main contributions are:
\begin{itemize}[leftmargin=*, label=$\bullet$]
    \item We propose \textbf{JOintGS}, a unified framework jointly optimizing camera trajectories, human poses, and 3D Gaussians from coarse initialization, enabling robust, calibration-free reconstruction and achieving SOTA performance.
    \item We introduce a \textbf{synergistic refinement mechanism} through explicit foreground-background disentanglement, where static backgrounds anchor camera optimization, refined cameras improve human body alignment, and optimized poses enhance scene reconstruction, forming a closed-loop of mutual reinforcement.
    \item We design efficient \textbf{temporal offset and residual modules} capturing fine-grained deformations and appearance variations while maintaining real-time rendering.  
\end{itemize}

%% file: figures/Overview_Framework.tex
\begin{figure}[t]
    \centering
    \includegraphics[width=\linewidth]{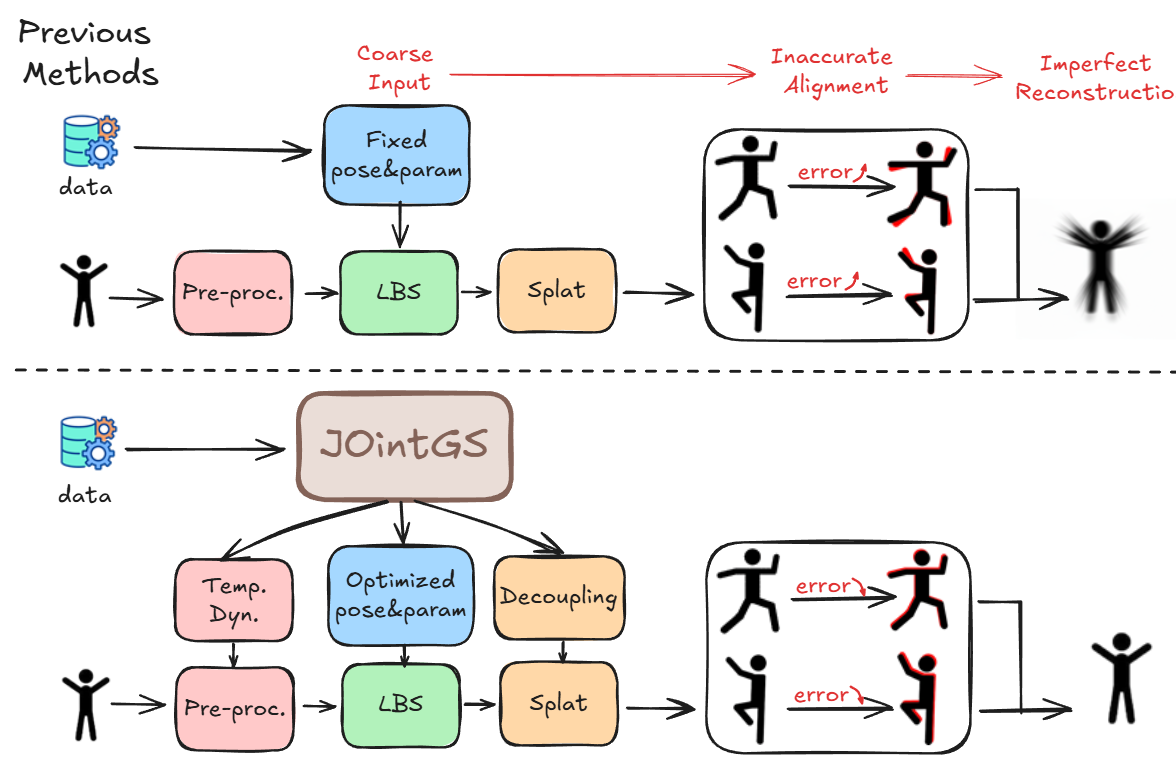}

    \caption{\textbf{Comparison with Previous Methods.} Unlike existing approaches that assume fixed camera poses and SMPL parameters as inputs, our JOintGS performs unified joint optimization through a synergistic refinement mechanism.
    }
    \label{fig:diff_others}
\end{figure}

%% file: sec/2_relatedworks.tex
\section{Related Work}
\label{sec:Related_Work}

\noindent\textbf{NeRF-based Human Reconstruction.} Neural radiance fields 
have enabled photorealistic human avatar reconstruction from monocular 
videos~\cite{weng2022humannerf,jiang2022neuman, li2024animatable, guo2023vid2avatar, peng2023implicit, feng2022capturing, mihajlovic2022keypointnerf, liu2021neural}. These methods achieve impressive quality by mapping posed observations to canonical 
space via SMPL-guided deformations~\cite{loper2023smpl}, but suffer from 
slow rendering due to expensive volume rendering. 
NeuMan~\cite{jiang2022neuman} jointly models humans and static scenes, while 
HumanNeRF~\cite{weng2022humannerf} focuses on human-only reconstruction. 
However, both assume fixed camera poses, limiting 
applicability to in-the-wild scenarios where pre-calibrated cameras are 
unavailable.

\noindent\textbf{3DGS-based Human Reconstruction.} Recent advances in 3D 
Gaussian Splatting (3DGS)~\cite{kerbl20233d} offer fast alternatives 
through explicit point-based representations. GaussianAvatar~\cite{hu2024gaussianavatar} binds 3D Gaussians to SMPL vertices with pose-dependent 
appearance modeling via hash-encoded MLPs, achieving real-time rendering. 3DGS-Avatar~\cite{qian20243dgs} explicitly models non-rigid 
deformations and pose-dependent color changes with MLPs, achieving high-fidelity reconstruction at the cost of 
increased training time. HUGS~\cite{kocabas2024hugs} jointly models humans 
and scenes using 3D Gaussians but lacks mechanisms for correcting initialization errors. Most existing methods critically 
depend on pre-calibrated cameras and accurate SMPL parameters as fixed inputs, 
which are difficult to obtain in real-world scenarios.

\noindent\textbf{Holistic Human-Scene Reconstruction.} Several methods 
attempt holistic reconstruction of both humans and scenes. Vid2Avatar~\cite{
guo2023vid2avatar} employs a human-centric scene model but struggles with 
proper multi-view geometry learning. HSR~\cite{xue2024hsr} extends 
Vid2Avatar with scene fields and holistic representation but shows degraded 
performance in outdoor scenes. ODHSR~\cite{zhang2025odhsr} proposes an 
online dense reconstruction framework, achieving impressive results through 
monocular geometric priors. However, their online updating strategy processes 
frames sequentially, potentially missing global optimization opportunities 
available when the full sequence is accessible.

\noindent\textbf{Joint Camera and Pose Optimization.} Traditional bundle 
adjustment~\cite{triggs1999bundle} jointly optimizes camera poses and 3D 
structure but struggles with dynamic scenes. Recent learning-based methods 
incorporate human priors into structure-from-motion. HSfM~\cite{muller2025reconstructing} integrates human reconstruction into classic SfM, 
demonstrating that modeling humans improves camera pose accuracy. 
PoseDiffusion~\cite{wang2023posediffusion} uses diffusion-aided bundle 
adjustment for pose estimation. Unlike these methods that operate on sparse 
features or separate optimization stages, \textit{our approach performs 
synergistic refinement within a unified differentiable rendering framework}, 
where dense photometric constraints from 3DGS enable tighter coupling between 
camera, pose, and appearance optimization.

%% file: sec/3_method.tex
\section{Method}
\label{sec:Method}

\subsection{Overview}
\label{subsec:overview}

Given a monocular RGB video $\{\mathbf{I}_t\}_{t=1}^T$ with coarse camera poses $\{\hat{\mathbf{T}}_t=[\hat{\boldsymbol{R}}|\hat{t}]\}$ from COLMAP~\cite{schoenberger2016sfm,schoenberger2016mvs} and initial SMPL parameters $\{\hat{\boldsymbol{\xi}}_t=(\hat{\boldsymbol{\theta}}_t, \hat{\boldsymbol{\beta}})\}$ from HMR2.0~\cite{zhang2023pymaf,cai2023smpler, goel2023humans}, our goal is to reconstruct a high-fidelity 
animatable human avatar and scene while simultaneously refining the camera pose and SMPL parameters.
\input{figures/JOintGS_Framework}
As illustrated in Figure~\ref{fig:JOintGS_Framework}, our joint optimization framework consists of four key components: (1)~\textit{Foreground human representation} (\S\ref{subsec:human_rep}) that models the avatar in canonical space with pose-driven deformation and temporal dynamics; (2)~\textit{Background scene representation} (\S\ref{subsec:background_rep}) using static 3D Gaussians; (3)~\textit{Synergistic refinement mechanism} (\S\ref{subsec:synergistic}) that simultaneously refines cameras, SMPL parameters, and Gaussian fields through unified differentiable rendering supervision.

\subsection{3D Human Representation}
\label{subsec:human_rep}

To model the dynamic human body with temporal variations, we represent the avatar as a collection of 3D Gaussians $\mathcal{G}_H = \{\mathbf{g}_i^H\}_{i=1}^{N_H}$ defined in a canonical space and deformed to arbitrary poses via learned skinning.

\subsubsection{Canonical Gaussian Field}
Following recent advances in Gaussian-based avatars~\cite{kocabas2024hugs,hu2024gauhuman}, we establish a \textit{canonical space} corresponding to SMPL rest pose (\eg, A-pose). Each human Gaussian $\mathbf{g}_i^H$ is parameterized by its canonical attributes: center position $\boldsymbol{\mu}_i^c \in \mathbb{R}^3$, rotation $\mathbf{R}_i^c \in SO(3)$, scale $\mathbf{S}_i^c \in \mathbb{R}^3_+$, opacity $\alpha_i \in [0,1]$, SH coefficients $\mathbf{c}_i$, and learned LBS weights $\mathbf{w}_i \in \mathbb{R}^K$. These canonical attributes are decoded from features sampled on a Triplane representation. 
We initialize $\mathcal{G}_H$ by uniformly sampling $N_H{=}110$k points on the SMPL mesh in the canonical pose.  
Each Gaussian inherits LBS weights $\mathbf{w}_i \in \mathbb{R}^K$ from its nearest SMPL vertex via barycentric interpolation, where $K{=}24$ denotes the number of joints.

\subsubsection{Pose-Driven Deformation}
To render canonical Gaussians under pose $\boldsymbol{\theta}_t$ at frame $t$, we apply standard LBS deformation~\cite{jung2023deformable, li2024animatable, pang2024ash}. Given SMPL parameters $(\boldsymbol{\theta}_t, \boldsymbol{\beta})$ and per-joint transformation matrices $\{\mathcal{T}_k^t \in \mathrm{SE}(3)\}_{k=1}^K$ computed via forward kinematics, we transform each Gaussian's attributes as:
\begin{align}
\boldsymbol{\mu}_i^t &= \sum_{k=1}^{K} w_{i,k} \, \mathcal{T}_k(\boldsymbol{\theta}_t) \, \begin{bmatrix} \boldsymbol{\mu}_i^c \\ 1 \end{bmatrix}, \label{eq:lbs_position} \\
\mathbf{R}_i^t &= \left(\sum_{k=1}^{K} w_{i,k} \, \mathbf{R}_k(\boldsymbol{\theta}_t)\right) \mathbf{R}_i^c, \label{eq:lbs_rotation}
\end{align}
where $\begin{bmatrix} \boldsymbol{\mu}_i^c;1 \end{bmatrix}$ denotes homogeneous coordinates and $\mathbf{R}_k(\boldsymbol{\theta}_t) = \mathcal{T}_k(\boldsymbol{\theta}_t)_{:3,:3}$ extracts the rotation component. We blend only rotational components for orientation to maintain shape consistency~\cite{qian2024gaussianavatars}. The covariance is updated as $\boldsymbol{\Sigma}_i^t = \mathbf{R}_i^t \, \mathbf{S}_i^c (\mathbf{S}_i^c)^\top \, (\mathbf{R}_i^t)^\top$.

\subsubsection{Temporal Dynamics Modeling}
While LBS-based deformation handles skeletal articulation, it cannot capture \textit{non-rigid dynamics} such as clothing wrinkles and \textit{appearance variations} caused by motions and lighting. To address this limitation, we introduce two lightweight modules that model residual deformations and appearance changes beyond the skeletal motion:

\noindent\textbf{Temporal Offset Module.} We model per-frame non-rigid geometric deformations via a shallow MLP $\mathcal{F}_{\text{offset}}$. It predicts positional and rotational offsets conditioned on the canonical position $\boldsymbol{\mu}_i^c$ and the encoded frame index $t$:
\begin{equation}
[\Delta \boldsymbol{\mu}_i^t, \Delta \mathbf{R}_i^t] = \mathcal{F}_{\text{offset}}(\mathcal{E}(\boldsymbol{\mu}_i^c), \gamma(t)),
\label{eq:offset_module}
\end{equation}
where $\mathcal{E}(\cdot)$ denotes multi-resolution hash encoding~\cite{muller2022instant} and $\gamma(\cdot)$ denotes positional encoding~\cite{vaswani2017attention}. The final posed position and rotation incorporate these offsets by:
\begin{equation}
\boldsymbol{\mu}_i^t \gets \boldsymbol{\mu}_i^t + \Delta \boldsymbol{\mu}_i^t, \quad \mathbf{R}_i^t \gets \mathbf{R}_i^{t} \cdot \Delta \mathbf{R}_i^t.
\label{eq:apply_offset}
\end{equation}

\noindent\textbf{Temporal Color Module.} To model temporal appearance variations (\eg, shadows, lighting changes), we predict a per-Gaussian color residual via another MLP $\mathcal{F}_{\text{color}}$, 
which is added to the base spherical harmonics color during rendering:
\begin{equation}
\mathbf{c}_i^t(\boldsymbol{\mu}_i^c) = \mathbf{c}_i^{\text{SH}}(\boldsymbol{\mu}_i^c) + 
\mathcal{F}_{\text{color}}(\mathcal{E}(\boldsymbol{\mu}_i^c), \gamma(t)).
\label{eq:color}
\end{equation}
These modules are regularized to remain small (see \S\ref{subsec:training_objectives}), ensuring they capture only the residual dynamics beyond skeletal motion.

\subsection{Background Scene Representation}
\label{subsec:background_rep}
 We represent the static 
scene as a set of 3D Gaussians, denoted as $\mathcal{G}_B = \{\mathbf{g}_i^B\}_{i=1}^{N_B}$, where each 
Gaussian is characterized by a center position $\boldsymbol{\mu}_i \in \mathbb{R}^3$, 
a 3D covariance matrix $\boldsymbol{\Sigma}_i$, an opacity $\alpha_i \in [0,1]$, 
and a view-dependent color parameterized by SH coefficients. We initialize $\mathcal{G}_B$ from the sparse point cloud reconstructed by COLMAP, which typically contains between $10\mathrm{k}$ and $50\mathrm{k}$ points, depending on the scene complexity. The initial Gaussian scales are set proportionally to the local point density, while their colors are inherited from the nearest image observations.

\subsection{Synergistic Refinement Mechanism}
\label{subsec:synergistic}

The core innovation of our method lies in the \textit{synergistic refinement} of three interdependent components, camera poses $\{\mathbf{T}_t\}$, SMPL parameters $\{(\boldsymbol{\theta}_t, \boldsymbol{\beta})\}$, and Gaussian fields $\{\mathcal{G}_H, \mathcal{G}_B\}$, within a unified differentiable rendering framework. 
Unlike prior works that optimize these components separately~\cite{qian2024gaussianavatars} or assume pre-calibrated inputs~\cite{jiang2023instantavatar,xu2023gaussianheadavatar}, our approach exploits their mutual dependencies through a closed-loop refinement process where each component progressively corrects errors in the others. 
To establish a unified world coordinate system, we employ RANSAC~\cite{fischler1981random} to fit scale-shift parameters, aligning SMPL depths to COLMAP's metric scale following.

This synergy operates through three complementary pathways that form a closed-loop of mutual reinforcement:

\noindent\textbf{Background-Anchored Camera Refinement.} Static background Gaussians $\mathcal{G}_B$ provide reliable multi-view geometric constraints across frames. Given initial camera pose $\hat{\mathbf{T}}_t$ from COLMAP, we parameterize a learnable correction:
\begin{equation}
\mathbf{T}_t = \Delta \mathbf{T}_t \circ \hat{\mathbf{T}}_t,
\label{eq:camera_refine}
\end{equation}
where $\Delta \mathbf{T}_t \in \mathrm{SE}(3)$ is represented via axis-angle and translation. We optimize $\Delta \mathbf{T}_t$ by minimizing photometric errors on static regions identified using human masks $\mathbf{M}_t$ from SAM~\cite{kirillov2023segment, ravi2024sam}:
\begin{equation}
\mathcal{L}_{\text{B}} = \|(1 - \mathbf{M}_t) \odot (\mathbf{I}_t - \hat{\mathbf{I}}_t^B)\|_1,
\label{eq:bg_loss}
\end{equation}
where $\hat{\mathbf{I}}_t^B$ is rendered from $\mathcal{G}_B$ and $\odot$ denotes element-wise multiplication. By isolating camera optimization from dynamic human motion, this strategy prevents artifacts from violating temporal consistency constraints.

\noindent\textbf{Camera-Guided Human Pose Optimization.} With refined cameras establishing accurate spatial-temporal correspondences, we optimize SMPL parameters $\{(\boldsymbol{\theta}_t, \boldsymbol{\beta})\}$ through photometric and silhouette supervision:
\begin{equation}
\mathcal{L}_{\text{H}} = \|\mathbf{M}_t \odot (\mathbf{I}_t - \hat{\mathbf{I}}_t^H)\|_1 + \|\mathbf{M}_t - \hat{\mathbf{M}}_t^H\|_1,
\label{eq:fg_loss}
\end{equation}
where $\hat{\mathbf{I}}_t^H$ and $\hat{\mathbf{M}}_t^H$ are rendered from $\mathcal{G}_H$. Improved camera alignment enables gradients from these losses to flow directly to SMPL parameters via the differentiable LBS transformation (Eq.~\ref{eq:lbs_position}--\ref{eq:lbs_rotation}), correcting initialization errors from monocular pose estimators.

\noindent\textbf{Pose-Aware Gaussian Optimization.} Refined camera and SMPL parameters enhance foreground-background disentanglement by providing accurate skeletal priors. This reduces foreground leakage into $\mathcal{G}_B$ and eliminates background artifacts from $\mathcal{G}_H$. We optimize Gaussian parameters using photometric losses:
\begin{equation}
\mathcal{L}_{\text{render}} = \lambda_{\text{rgb}} \mathcal{L}_{\text{rgb}} + \lambda_{\text{ssim}} \mathcal{L}_{\text{ssim}} + \lambda_{\text{lpips}} \mathcal{L}_{\text{lpips}},
\label{eq:render_loss}
\end{equation}
where $\mathcal{L}_{\text{rgb}} = \|\mathbf{I}_t - \hat{\mathbf{I}}_t\|_1$ is $\ell_1$ loss, $\mathcal{L}_{\text{ssim}}$~\cite{wang2004image} measures structural similarity, and $\mathcal{L}_{\text{lpips}}$~\cite{zhang2018unreasonable} captures perceptual quality. Cleaner scene separation further stabilizes camera estimation by removing dynamic motion from static constraints, completing the feedback loop.

\subsection{Training Strategy and Objectives}
\label{subsec:training_objectives}

We employ a structured \textbf{three-stage} optimization schedule to prevent degenerate solutions and ensure stable convergence. 
\emph{(i) Warm-up:} We optimize only Gaussian parameters $\{\mathcal{G}_H, \mathcal{G}_B\}$ with fixed camera and SMPL parameters, establishing a reliable geometry prior before moving to pose-related updates. 
\emph{(ii) Independent Optimization:} We simultaneously enable the optimization of camera poses and SMPL parameters $\{(\boldsymbol{\theta}_t, \boldsymbol{\beta})\}$. During this stage, these components are updated independently to avoid gradient interference: cameras are primarily anchored by static background cues for multi-view consistency, while SMPL parameters are refined to correct pose initialization errors based on human-centric gradients. 
\emph{(iii) Joint Optimization:} We perform full optimization with complete losses, allowing for \textbf{synergistic refinement} where camera trajectories, body poses, and Gaussian attributes mutually correct each other to achieve global consistency. 
More details are provided in the supplementary material.

To prevent overfitting and maintain generalization, we introduce extra complementary regularizations. \emph{LBS weight regularization} constrains learned weights to remain close to SMPL initialization:
\begin{equation}
    \mathcal{L}_{\text{lbs}} = \sum_{i=1}^{N_H} \|\mathbf{w}_i - \mathbf{w}_i^{\text{SMPL}}\|_2^2,
\end{equation}
preventing skinning weights from overfitting to training poses. \emph{Offset regularization} penalizes large deformations to ensure temporal modules capture only residual dynamics:
\begin{equation}
    \mathcal{L}_{\text{offset}} = \sum_{i=1}^{N_H} \sum_{t=1}^{T} (\|\Delta \boldsymbol{\mu}_i^t\|_2^2 + \|\Delta \mathbf{R}_i^t - \mathbf{I}\|_F^2 + \|\Delta \mathbf{c}_i^t\|_2^2),
\end{equation}
where $\mathbf{I}$ is the identity rotation. \emph{Canonical regularization} softly anchors human Gaussians near SMPL mesh surface:
\begin{equation}
    \mathcal{L}_{\text{canonical}} = \sum_{i=1}^{N_H} \min_{j \in \{1, \dots, N_v\}} \|\boldsymbol{\mu}_i^c - \bar{\mathbf{v}}_j\|_2^2, 
\end{equation}
where $\{\bar{\mathbf{v}}_j\}$ are canonical SMPL vertices.

%% file: figures/JOintGS_Framework.tex
\begin{figure*}[t]
    \centering
    \includegraphics[width=0.9\linewidth]{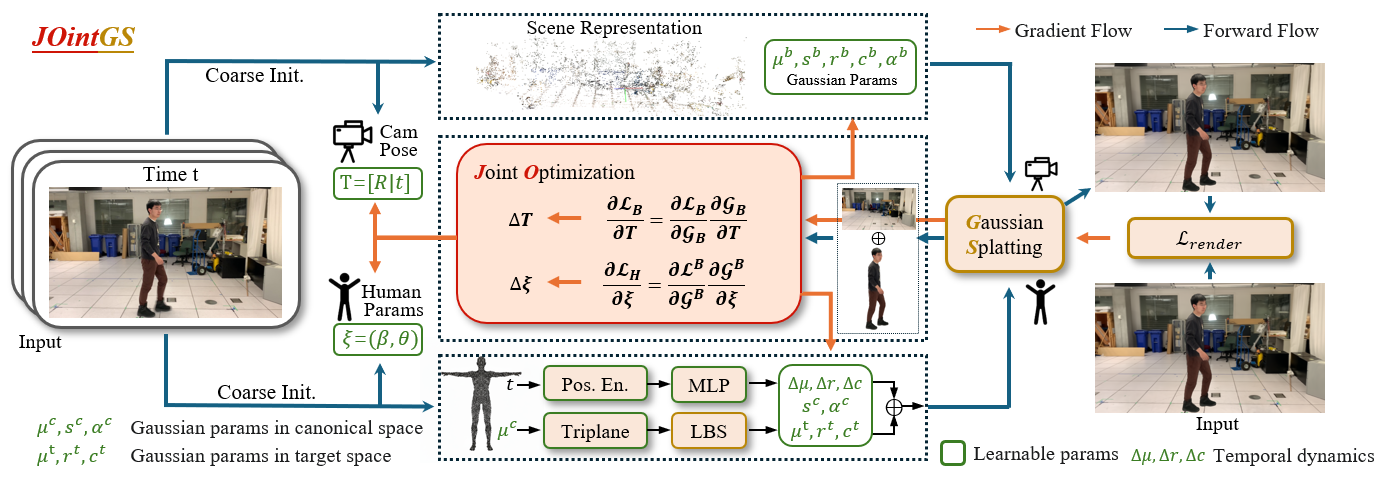}
    \caption{\textbf{JOintGS Framework Overview.} Given a monocular RGB video with coarse camera poses $\boldsymbol{T}=[\boldsymbol{R}|\mathbf{t}]$ from COLMAP and initial SMPL parameters $\boldsymbol{\xi}=(\boldsymbol{\beta},\boldsymbol{\theta})$ from HMR2.0, we initialize scene Gaussians $\mathcal{G}_B$ (from COLMAP point cloud) and human Gaussians $\mathcal{G}_H$ (from SMPL vertices) in canonical space.
    Our synergistic refinement mechanism (highlighted by orange gradient flow) jointly optimizes camera pose corrections $\Delta\boldsymbol{T}$, SMPL parameter refinements $\Delta\boldsymbol{\xi}$, and Gaussian attributes $\{\mathcal{G}_H, \mathcal{G}_B\}$ through unified differentiable rendering supervision. The optimization operates through three complementary pathways: (1) \textbf{Background-anchored camera refinement:} static scene Gaussians provide multi-view geometric constraints via photometric loss $\mathcal{L}_B$ on background regions; (2) \textbf{Camera-guided human optimization:} refined cameras enable accurate temporal correspondence for SMPL parameter optimization via human rendering loss $\mathcal{L}_H$; (3) \textbf{Pose-aware Gaussian optimization:} improved camera and SMPL parameters enhance foreground-background disentanglement, facilitating Gaussian field optimization with photometric losses $\mathcal{L}_{\text{render}}$. This closed-loop mutual refinement enables robust reconstruction from noisy initialization without requiring pre-calibrated inputs.
    }
    \label{fig:JOintGS_Framework}
\end{figure*}

%% file: sec/4_experiments.tex
\section{Experiments}
\label{sec:Experiments}

We conduct comprehensive experiments to evaluate the effectiveness of our 
proposed method. We first describe the datasets and evaluation metrics 
(§\ref{sec:Dataset}), followed by implementation details (§\ref{sec:implementation}). 
We then present quantitative (§\ref{sec:quantitative}) and qualitative 
(§\ref{sec:qualitative}) comparisons with state-of-the-art methods. Finally, ablation 
studies validate the design choices of our key components (§\ref{sec:ablation}).

\subsection{Dataset}
\label{sec:Dataset}
\input{tables/nv_neuman}
\input{tables/nv_EMDB}
\noindent\textbf{NeuMan Dataset}~\cite{jiang2022neuman} comprises six in-the-wild   
sequences (Seattle, Citron, Parking, Bike, Jogging, Lab), each capturing a single   
person performing various activities over 10–20 seconds. The videos are recorded   
with a handheld mobile phone exhibiting natural camera motion, which provides   
sufficient viewpoint diversity for multi-view reconstruction. We follow the   
original split protocol~\cite{jiang2022neuman}, allocating 80\% of frames for   
training, 10\% for validation, and 10\% for testing. Notably, we \textit{do not} use the keyframe selection strategy on this dataset to enable direct comparison   
with prior offline methods. This dataset serves as our primary benchmark for quantitative evaluation.  

\noindent\textbf{EMDB Dataset}~\cite{kaufmann2023emdb} is a large-scale   
in-the-wild dataset consisting of 81 video sequences from 10 subjects, totaling   
58 minutes of motion data. The dataset is captured using Wearable trackers and   
a handheld iPhone, providing ground-truth global camera poses and body root   
trajectories via wireless motion capture sensors. We select ten representative sequences that   
present diverse challenges: extended trajectories, occlusions, complex lighting   
(shadows), and unconventional poses (\eg, cartwheels). We use the first 200 frames from   
each sequence and adopt an 80\%/10\%/10\% train/val/test split.

\noindent\textbf{Baselines.} We compare against state-of-the-art methods across 
different categories: (1) \textit{NeRF-based human reconstruction}: 
HumanNeRF~\cite{weng2022humannerf}, InstantAvatar~\cite{jiang2023instantavatar}; 
(2) \textit{3DGS-based human reconstruction}: 3DGS-Avatar~\cite{qian20243dgs}, 
GaussianAvatar~\cite{hu2024gaussianavatar}, ExAvatar~\cite{moon2024expressive}; 
(3) \textit{Video-based human reconstruction}: Vid2Avatar~\cite{guo2023vid2avatar}, Vid2Avatar-Pro~\cite{guo2025vid2avatar}; 
(4) \textit{Holistic human-scene reconstruction}: NeuMan~\cite{jiang2022neuman}, 
HUGS~\cite{kocabas2024hugs}, HSR~\cite{xue2024hsr}, ODHSR~\cite{zhang2025odhsr}.

\noindent\textbf{Evaluation Metrics.} We evaluate reconstruction quality using 
standard photometric metrics: PSNR, SSIM~\cite{wang2004image}, and LPIPS~\cite{zhang2018unreasonable}. For human-only rendering, we composite the reconstructed avatar onto a white background and compute metrics 
over the entire image region. We also report training time (hours) and 
rendering speed (FPS) to assess computational efficiency.

\subsection{Implementation Details}
\label{sec:implementation}
All experiments were conducted on a single NVIDIA RTX 5090 GPU (32GB). The model was optimized using the Adam~\cite{kinga2015method} optimizer, with a stage-dependent learning rate scheduling strategy employed throughout the training process. Training converged in approximately 25 minutes over 15,000 iterations. Detailed loss weights, learning rate schedules, initialization methods, and network architectures are provided in the supplementary material.

\input{figures/qualitative}

\input{figures/novelty_pose}
\input{figures/robust}

\subsection{Quantitative Results}
\label{sec:quantitative}

\noindent\textbf{Novel View Synthesis on NeuMan.} Table~\ref{tab:neuman} presents quantitative evaluation results on the NeuMan dataset. Our method achieves superior performance across all metrics, an average PSNR of 34.84~dB, 
surpassing the previous best Vid2Avatar-Pro~\cite{guo2025vid2avatar} 
(32.71~dB) by 2.13~dB. Notably, 3DGS-based methods (3DGS-Avatar, GaussianAvatar, 
HUGS, Ours) consistently outperform NeRF-based approaches (HumanNeRF, 
InstantAvatar) in both quality and speed, demonstrating the advantages of 
explicit 3D Gaussian representation for human modeling. Our method also 
achieves highest SSIM (0.984) and LPIPS (0.010), indicating 
excellent structural similarity and perceptual quality.

\noindent\textbf{Novel View Synthesis on EMDB.} Table~\ref{tab:emdb} shows 
results on the more challenging EMDB dataset featuring complex poses and 
occlusions. Our method achieves 30.99~dB PSNR, surpassing the recent online 
method ODHSR~\cite{zhang2025odhsr} (28.95~dB) by 2.04~dB. 
Despite ODHSR's online advantage enabling real-time tracking, our offline 
global optimization leverages full sequence information for more robust 
reconstruction. The consistent improvements across both datasets validate the 
robustness and generalizability of our approach.

\noindent\textbf{Efficiency Analysis.} JointGS 
requires approximately 23 minutes for training on average, slightly 
faster than HUGS (25 minutes), while achieving significantly better quality 
(2.13~dB PSNR). Our method maintains real-time rendering at 27.3 FPS versus 
HUGS's 27.5 FPS. This efficiency is attributed to our lightweight temporal offset module and residual color field.

\noindent\textbf{Robustness to Initialization Errors.} Figure~\ref{fig:robust} 
analyzes the robustness of our method to noisy initialization of camera poses 
and SMPL parameters. We incrementally add Gaussian noise to the initialization 
with standard deviations ranging from 0 to 0.02 (normalized scale). Our method 
exhibits significantly slower performance degradation compared to HUGS: at 
$\sigma{=}0.01$, our PSNR drops by only 0.9~dB versus HUGS's 3.7~dB drop, 
demonstrating a 2.8~dB robustness advantage. This robustness is achieved 
through the synergistic refinement mechanism, which iteratively refines both 
camera and SMPL parameters within the joint optimization framework.

\noindent\textbf{Background Reconstruction Results.} 
While our background model primarily serves to provide correct context (\eg, occlusion and depth cues) and anchor camera refinement, it also yields high-quality scene reconstruction as a byproduct of joint optimization. As shown in Tables~1 and 2, our method surpasses the recent ODHSR by {2.45 dB} in PSNR on the NeuMan dataset and achieves comparable performance on EMDB. Per-scene breakdowns and visual comparisons are provided in the supplementary material.

\subsection{Qualitative Results}
\label{sec:qualitative}

\noindent\textbf{Human Reconstruction Quality.} Figure~\ref{fig:qualitative} 
presents qualitative comparisons of human reconstruction against NeuMan~\cite{
jiang2022neuman} and HUGS~\cite{kocabas2024hugs}. For each example, we show 
the full-body rendering (left) alongside two zoomed-in views highlighting 
fine-grained details. Our method demonstrates superior reconstruction quality 
across three key aspects:

\textit{Body Alignment.} JointGS accurately aligns the reconstructed body with 
the ground-truth pose, avoiding the misalignment artifacts visible in baseline 
methods. This is achieved through our synergistic optimization of camera poses, 
SMPL parameters, and Gaussian representations.

\textit{Detail Preservation.} The zoomed-in regions reveal that our method 
preserves fine-grained details such as facial features, clothing wrinkles, and 
hand gestures more faithfully than baselines. This benefit stems from the 
temporal offset module, which captures high-frequency deformations beyond 
SMPL's rigid skeletal deformations.

\textit{Color Fidelity.} Our reconstructions exhibit more accurate color 
reproduction compared to HUGS, which suffers from color bleeding between human 
and scene. The clean separation is enabled by our synergistic refinement 
mechanism, which refines SMPL parameters to reduce ambiguity in human-scene 
decomposition.

\noindent\textbf{Novel Pose and Environment Transfer.} As shown in 
Figure~\ref{fig:novelty_pose}, our method enables decoupling of human and 
background, facilitating novel view synthesis and avatar manipulation. We 
demonstrate this capability by transferring a reconstructed human avatar from 
the Lab sequence to the Parking  sequence environment, and rendering it under a 
novel pose extracted from a different subject. The results exhibit realistic 
appearance and consistent geometry, validating that JointGS effectively 
separates human from the background while maintaining animatable canonical 
representations.

\input{tables/ablation}
\input{figures/ablation}
\subsection{Ablation Experiments}
\label{sec:ablation}

We conduct ablation studies to validate the contribution of each proposed component on the NeuMan dataset (Table~\ref{tab:ablation} and Figure~\ref{fig:ablation}). 
Removing the \textit{temporal dynamics module} (w/o Dynamics) results in a $0.6\,\mathrm{dB}$ PSNR drop with visible degradation in high-frequency details such as clothing wrinkles and facial features, validating its necessity for capturing pose-dependent non-rigid deformations beyond SMPL's skeletal articulation. 
More critically, removing \textit{synergistic refinement} (w/o Synergistic)—which jointly optimizes cameras, SMPL parameters, and Gaussians—leads to a substantial $3.5\,\mathrm{dB}$ degradation with severe artifacts including misaligned limbs and blurred textures, demonstrating that correcting initialization errors through mutual reinforcement is essential for accurate reconstruction. 
Our full model achieves $34.84\,\mathrm{dB}$ PSNR on average, confirming that geometric refinement (temporal offsets), appearance modeling (residual colors), and holistic optimization (synergistic mechanism) work synergistically to enable high-fidelity reconstruction from coarse initialization.

%% file: tables/nv_neuman.tex
\begin{table*}[t]
\centering
\small
\setlength{\tabcolsep}{8pt}
\begin{tabular}{l|ccc|ccc}
\toprule
& \multicolumn{3}{c|}{\textbf{Human-only}} & \multicolumn{3}{c}{\textbf{Full-image}} \\
\textbf{Method} & \textbf{PSNR}$\uparrow$ & \textbf{SSIM}$\uparrow$ & \textbf{LPIPS}$\downarrow$ & \textbf{PSNR}$\uparrow$ & \textbf{SSIM}$\uparrow$ & \textbf{LPIPS}$\downarrow$ \\
\midrule
Vid2Avatar~\cite{guo2023vid2avatar}          & 30.96 & 0.981 & 0.018 & 15.64 & 0.551 & 0.572 \\
HUGS~\cite{kocabas2024hugs}                  & 30.13 & 0.977 & 0.017 & 26.66 & 0.851 & 0.126 \\
HSR~\cite{xue2024hsr}                        & 29.03 & 0.971 & 0.026 & 21.67 & 0.669 & 0.526 \\
ODHSR~\cite{zhang2025odhsr}                  & 32.07 & 0.981 & 0.016 & 27.78 & 0.870 & 0.153 \\
ExAvatar~\cite{moon2024expressive}           & 31.39 & 0.981 & 0.016 & - & - & - \\
Vid2Avatar-Pro~\cite{guo2025vid2avatar}      & 32.71 & 0.983 & 0.019 & - & - & - \\
\midrule
\textbf{JOintGS (Ours)} & \textbf{34.84} & \textbf{0.984} & \textbf{0.010} & \textbf{30.23} & \textbf{0.913} & \textbf{0.072} \\
\bottomrule
\end{tabular}
\caption{Quantitative evaluation on NeuMan dataset~\cite{jiang2022neuman}. We report performance on both human-only regions and entire frames (Full-image). For the \textbf{human-only} setting, we render the avatar on a white background for all baselines and compute metrics over the whole image.}
\label{tab:neuman}
\label{tab:neuman}
\end{table*}

%% file: tables/nv_EMDB.tex
\begin{table*}[t]
\centering
\small
\setlength{\tabcolsep}{8pt}
\begin{tabular}{l|ccc|ccc}
\toprule
& \multicolumn{3}{c|}{\textbf{Human-only}} & \multicolumn{3}{c}{\textbf{Full-image}} \\
\textbf{Method} & \textbf{PSNR}$\uparrow$ & \textbf{SSIM}$\uparrow$ & \textbf{LPIPS}$\downarrow$ & \textbf{PSNR}$\uparrow$ & \textbf{SSIM}$\uparrow$ & \textbf{LPIPS}$\downarrow$ \\
\midrule
GauHuman~\cite{hu2024gauhuman}          & 25.31 & 0.943 & 0.057 & - & - & - \\
3DGS-Avatar~\cite{qian20243dgs}    & 27.95 & 0.967 & 0.035 & - & - & - \\
Vid2Avatar~\cite{guo2023vid2avatar}     & 24.25 & 0.948 & 0.061 & 16.65 & 0.413 & 0.599 \\
HUGS~\cite{kocabas2024hugs}             & 26.16 & 0.947 & 0.033 & 21.60 & 0.659 & 0.181 \\
HSR~\cite{xue2024hsr}                   & 25.12 & 0.920 & 0.054 & 18.67 & 0.463 & 0.632 \\
ODHSR~\cite{zhang2025odhsr}             & 28.95 & 0.966 & 0.031 & \textbf{23.79} & 0.767 & 0.197 \\
\midrule
\textbf{JOintGS (Ours)} & \textbf{30.99} & \textbf{0.972} & \textbf{0.027} & 23.40 & \textbf{0.785} & \textbf{0.173} \\
\bottomrule
\end{tabular}
\caption{Quantitative evaluation on EMDB dataset~\cite{kaufmann2023emdb}. We report metrics on both human-only regions and entire frames (Full-image) to provide a multi-faceted assessment. Our method outperforms both offline and online baselines.}
\label{tab:emdb}
\end{table*}

%% file: figures/qualitative.tex
\begin{figure*}[t]
    \centering
    \includegraphics[width=0.9\linewidth]{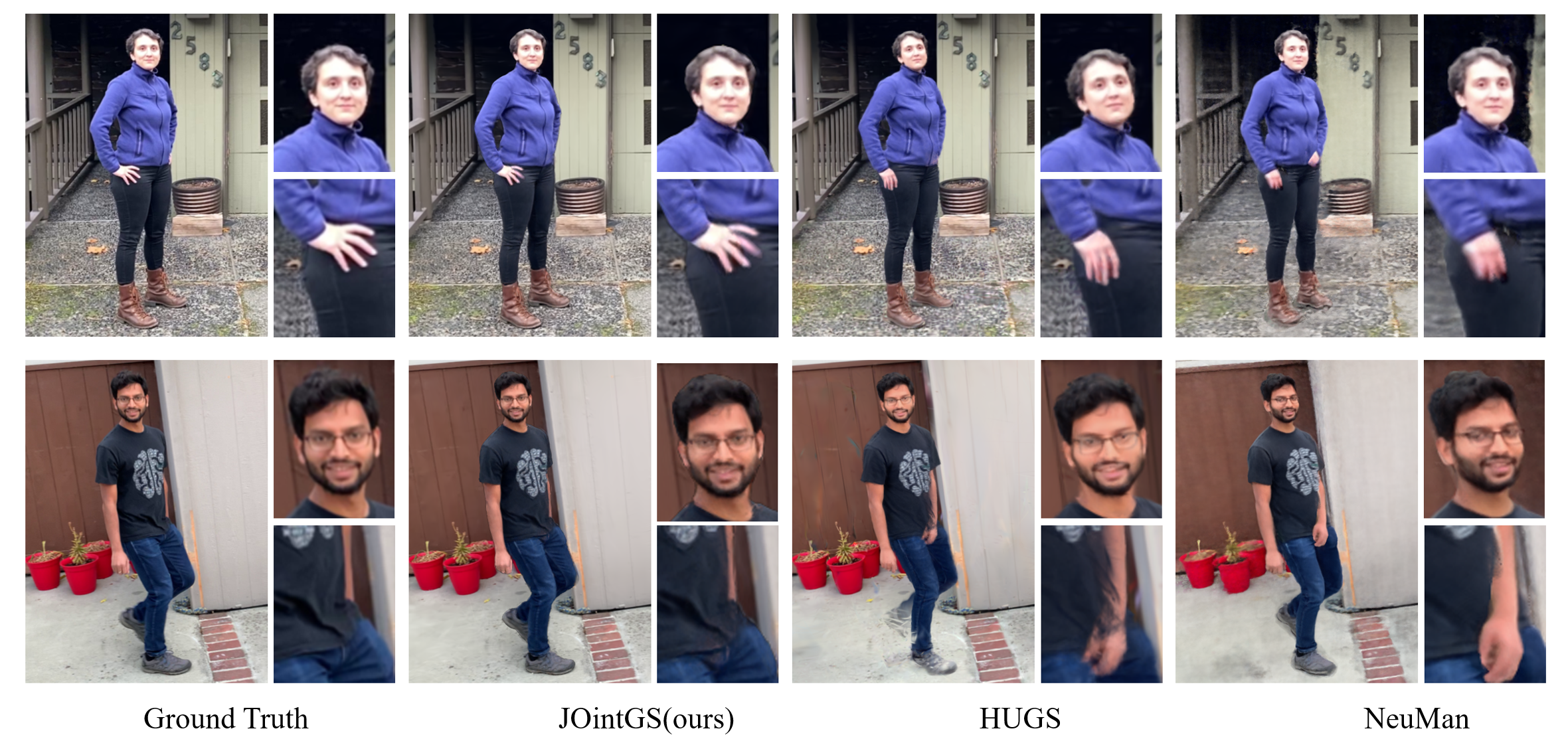}
    \caption{\textbf{Qualitative comparison on NeuMan dataset.} For each scene, we show full-body rendering (left) and zoomed-in details (right).} 
    \label{fig:qualitative}
\end{figure*}

%% file: figures/novelty_pose.tex
\begin{figure}[t]
    \centering
    \includegraphics[width=0.9\linewidth]{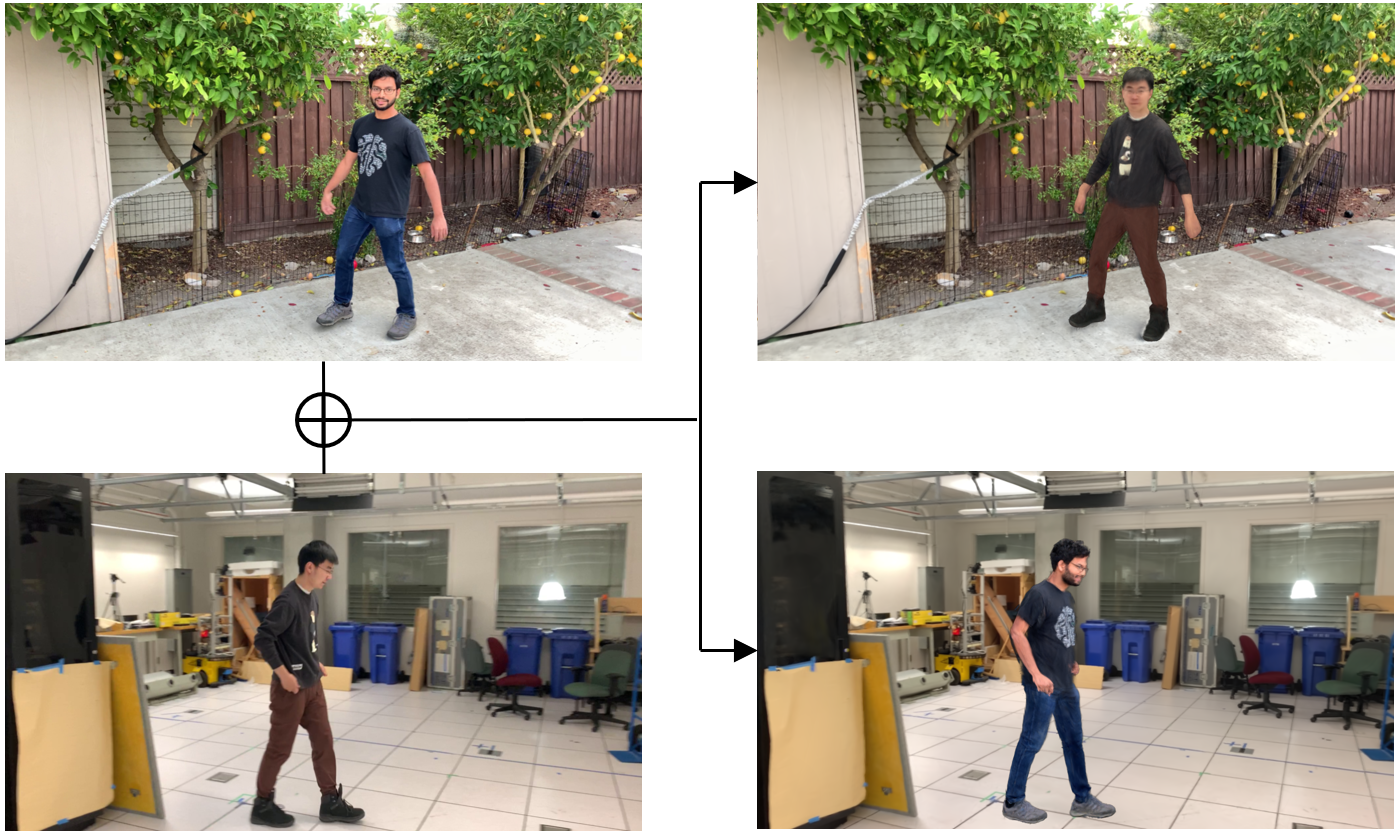}
    \caption{\textbf{Novel Pose Synthesis And Environment Transfer.} The reconstructed avatar can be animated with arbitrary poses while maintaining photorealistic appearance and seamless integration with other environments.}
    \label{fig:novelty_pose}
\end{figure}

%% file: figures/robust.tex
\begin{figure}[t]
    \centering
    \includegraphics[width=0.9\linewidth]{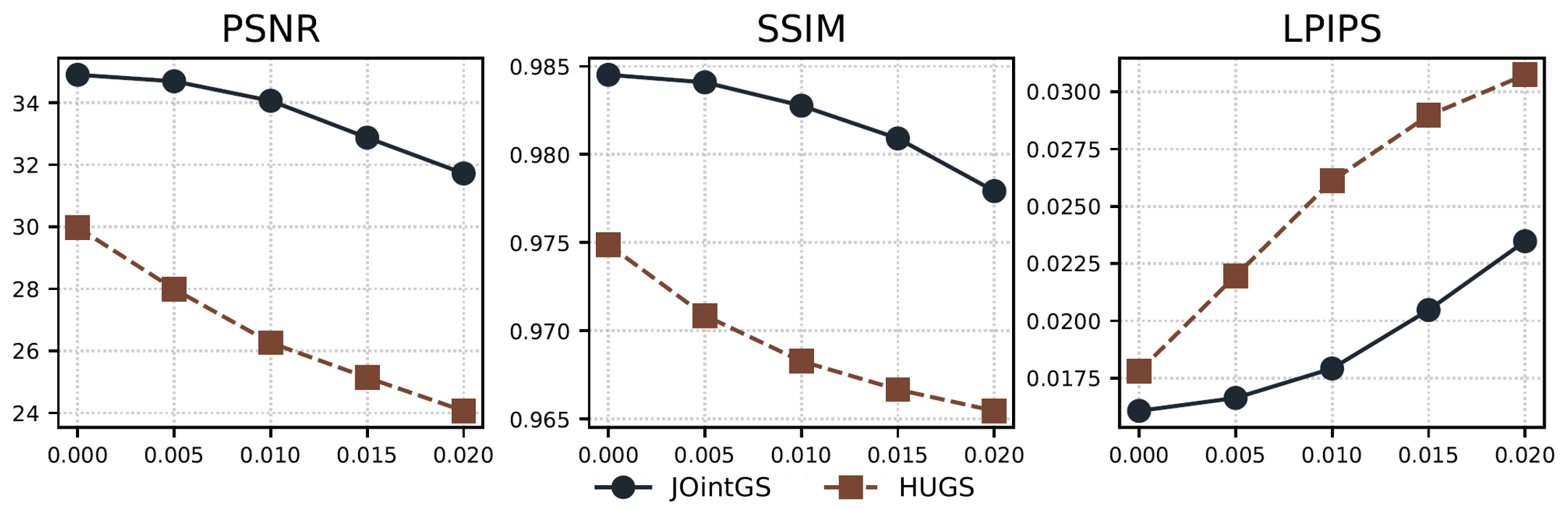}
    \caption{\textbf{Robustness Comparison } on noisy initialization.} 
    \label{fig:robust}
\end{figure}

%% file: tables/ablation.tex
\begin{table}[t]
\centering

\resizebox{0.85\columnwidth}{!}{
\begin{tabular}{l|ccc}
\toprule
\textbf{Configuration} & \textbf{PSNR}$\uparrow$ & \textbf{SSIM}$\uparrow$ & 
\textbf{LPIPS}$\downarrow$ \\
\midrule
w/o Synergistic &  31.38&  0.976&  0.017  \\
w/o Dynamics &  34.23&  0.983&  0.011  \\
\midrule
\textbf{Full Model (Ours)} & \textbf{34.84} & \textbf{0.984} & \textbf{0.010} \\
\bottomrule
\end{tabular}
}
\caption{Ablation study on NeuMan dataset (average across six sequences). Each 
component contributes to the final performance, with synergistic refinement 
providing the largest gain.}
\label{tab:ablation}
\end{table}

%% file: figures/ablation.tex
\begin{figure}[t]
    \centering
    \includegraphics[width=\linewidth]{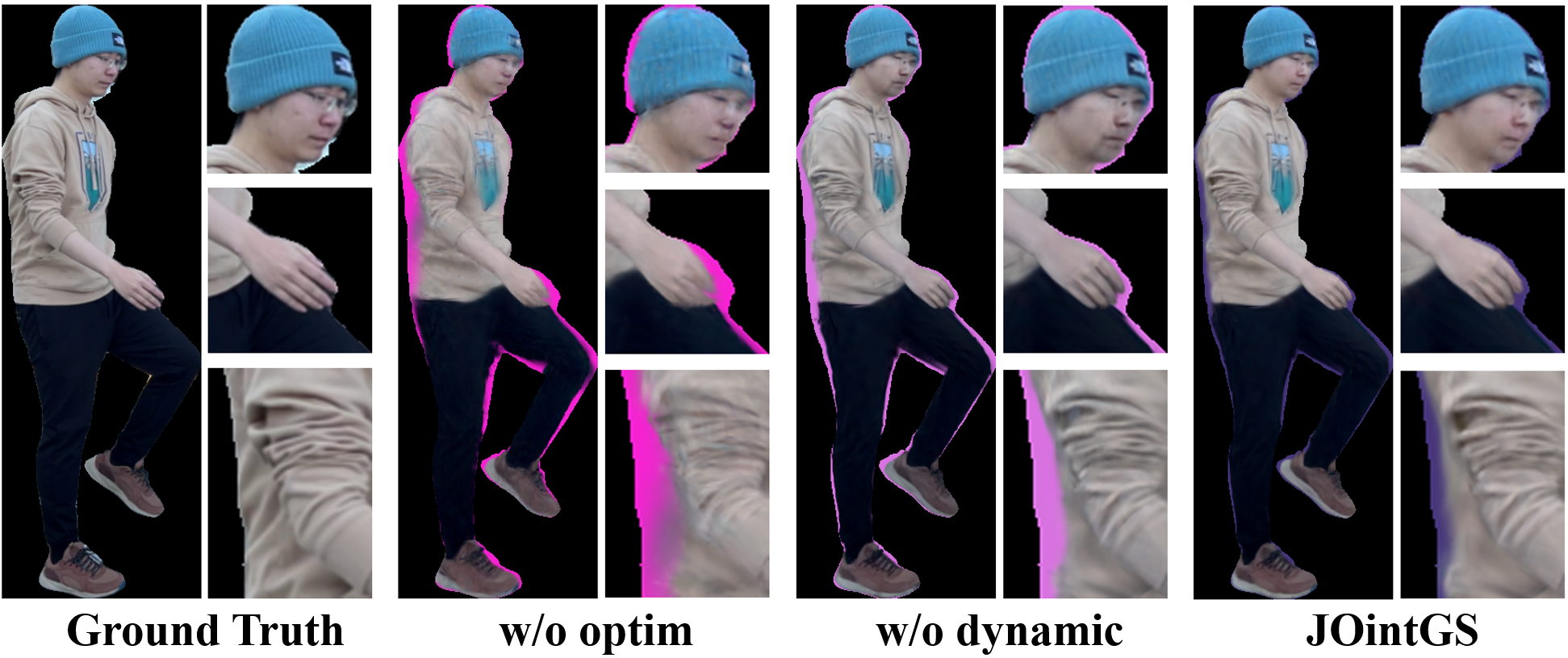}
    \caption{Qualitative visualization of the details captured in the human body reconstruction under different ablations of our method.} 
    \label{fig:ablation}
\end{figure}

%% file: sec/5_conclusion.tex
\section{Conclusion}  
\label{sec:Conclusion}  

We present \textbf{JointGS}, a unified framework for holistic reconstruction of dynamic humans and static scenes from monocular RGB videos through synergistic refinement of camera poses, SMPL parameters, and 3D Gaussian fields. By exploiting their mutual dependencies within a differentiable rendering pipeline, our method enables robust reconstruction from coarse initialization without demanding precise pre-calibration. Extensive experiments on diverse in-the-wild datasets demonstrate that JointGS achieves state-of-the-art performance, while maintaining real-time rendering. Comprehensive ablation studies validate the effectiveness of each proposed component, particularly the synergistic refinement mechanism.

\noindent\textbf{Limitations.} Our approach is fundamentally constrained by the inherent capacity of the SMPL body model, leading to reduced fidelity in fine-grained regions such as hands and faces. Although our temporal dynamics module exhibits a tendency to capture residual detail in these areas, a promising future direction is to improve hand and face modeling and enhance the controllability of expressions and gestures by integrating more expressive parametric models~\cite{shen2023xavatar}, such as SMPL-X~\cite{pavlakos2019expressive}.

\section*{Impact Statement}
This paper presents work whose goal is to advance the field of Machine Learning. There are many potential societal consequences of our work, none of which we feel must be specifically highlighted here.

%% file: sec/X_suppl.tex
\clearpage
\setcounter{page}{1}

\section{Overview}
This supplementary material provides further details regarding our proposed model, JOintGS. These details were omitted from the main paper due to space constraints. Furthermore, we include additional experimental results demonstrating that our model achieves State-of-the-Art (SOTA) performance in background reconstruction, even though the background was not the primary target of our optimization. 

The supplementary material is organized as follows:

\begin{itemize}
    \item Section\ref{sec:Supplementary_Implementation}: Detailed network architectures, advanced training schemes, and the loss function formulation.
    \item Section\ref{sec:Supplementary_Experiments}: Additional Experimental Results and Additional Demonstration Video.
\end{itemize}

\section{Implementation Details}
\label{sec:Supplementary_Implementation}

\subsection{Network Architectures}

We present a lightweight Multi-Layer Perceptron (MLP) network for predicting individual Gaussian properties, as illustrated in Figure~\ref{fig:model_architecture}.   

Following HUGS~\cite{kocabas2024hugs}, we employ a Tri-Plane encoding to extract spatial features for each Gaussian, which serve as input to our MLP-based decoder networks. Specifically, for a Gaussian positioned at $\mathbf{p} \in \mathbb{R}^3$, we first project it onto three orthogonal planes (XY, YZ, XZ), obtaining corresponding 2D coordinates: $\mathbf{p}_{xy}, \mathbf{p}_{yz}, \mathbf{p}_{xz}$. Utilizing bilinear interpolation, we sample the triplane feature maps at these coordinates, yielding three feature vectors: $\mathbf{f}_{xy}, \mathbf{f}_{yz}, \mathbf{f}_{xz}$. These vectors are concatenated to form the final sampling strategyspatial feature vector $\mathbf{f}_{\text{spatial}} = [\mathbf{f}_{xy}; \mathbf{f}_{yz}; \mathbf{f}_{xz}]$. To capture the temporal dynamics of human motion, we further augment the spatial feature with a time embedding, obtained by applying a positional encoding function to the time step $t$. Consequently, the final input feature for each Gaussian is $\mathbf{f}_{\text{input}} = [\mathbf{f}_{\text{spatial}}; \mathbf{f}_{\text{time}}]$.  

We design separate Appearance and Geometry Decoders to predict distinct Gaussian attributes. The Geometry Decoder is responsible for estimating spatial properties, including mean position ($\mu$), rotation ($r$), and scale ($s$). Conversely, the Appearance Decoder focuses on predicting visual attributes, namely color ($c$) and opacity ($o$). This architectural disentanglement enables more specialized learning and enhanced representation of geometric and appearance features. 

\input{figures/ablation_stage}
\input{figures/model_architecture}

\subsection{Training Details}
As introduced in the main paper (Sec. 3.5), we employ a carefully designed four-stage optimization schedule. During the Warm-up Stage, the initial learning rate is strategically set to a relatively high value. Specifically, the Gaussian attributes of the background component ($\mu, o, s, c, a$) are assigned learning rates of $1.6 \times 10^{-4}$, $0.05$, $0.005$, $0.001$, and $0.0025$, respectively, while the foreground position ($\mu$) and the triplane network receive learning rates of $1.6 \times 10^{-4}$ and $0.001$. This stage proceeds for 5,000 iterations. For computational efficiency, the Camera Optimization and Human Optimization proceed concurrently. In this parallel process, the camera pose only receives gradients originating from the background Gaussians, and the human parameters exclusively receive gradients from the foreground Gaussians. Both the camera pose and human parameters are optimized with a learning rate of $0.001$ for 5,000 iterations. 
To verify the necessity of this phased design, we provide an additional ablation study and visualize the results in Figure~\ref{fig:ablation_stage}. The results demonstrate that skipping the initial stages and proceeding directly to joint optimization leads to severe gradient interference; specifically, a significant amount of foreground-related information is incorrectly captured by the background Gaussians.
Finally, in the Joint Optimization Stage, we remove all constraints on the gradient graph, allowing the camera pose and human parameters to participate in synergistic refinement with both foreground and background Gaussians, thereby achieving the mutual correction of each component. In this stage, learning rates follow a cosine annealing schedule, decaying to $0.1$ times their initial value after 10,000 iterations.

\subsection{Loss Function Formulation}
The overall loss is a weighted sum of three major components: rendering loss ($\mathcal{L}_{\text{render}}$), prior loss ($\mathcal{L}_{\text{prior}}$), and regularization loss ($\mathcal{L}_{\text{regular}}$):
\begin{equation}
    \mathcal{L} = \mathcal{L}_{\text{render}} + \mathcal{L}_{\text{prior}} + \mathcal{L}_{\text{regular}}
\end{equation}

\subsubsection{Rendering Loss}
The rendering loss, $\mathcal{L}_{\text{render}}$, represents the pixel-level fidelity between the rendered image and the ground truth. It is a combination of three metrics: $\mathcal{L}_{\text{rgb}}$ (pixel-wise $\mathcal{L}_1$ loss), $\mathcal{L}_{\text{ssim}}$ (Structural Similarity Index), and $\mathcal{L}_{\text{lpips}}$ (Learned Perceptual Image Patch Similarity):
\begin{equation}
    \mathcal{L}_{\text{render}} = \lambda_{\text{rgb}} \mathcal{L}_{\text{rgb}} + \lambda_{\text{ssim}} \mathcal{L}_{\text{ssim}} + \lambda_{\text{lpips}} \mathcal{L}_{\text{lpips}}
\end{equation}
We set the corresponding weights as $\lambda_{\text{rgb}} = 1$, $\lambda_{\text{ssim}} = 0.4$, and $\lambda_{\text{lpips}} = 0.2$.

\subsubsection{Prior Loss}
The prior loss, $\mathcal{L}_{\text{prior}}$, incorporates prior knowledge into the optimization process, specifically ensuring the rendered human avatar adheres to a segmentation mask:
\begin{equation}
    \mathcal{L}_{\text{prior}} = \lambda_{\text{mask}} \mathcal{L}_{\text{mask}}
\end{equation}
Here, $\mathcal{L}_{\text{mask}}$ is the Mean Squared Error (MSE) loss between the rendered human body segmentation and the provided prior mask. The weight for this term is $\lambda_{\text{mask}} = 0.01$.

\subsubsection{Regularization Loss}
The regularization term, $\mathcal{L}_{\text{regular}}$, is introduced to maintain model stability and geometric plausibility, encompassing Linear Blend Skinning (LBS) constraints, geometric constraints with the SMPL model, and dynamic attribute regularization:
\begin{equation}
    \mathcal{L}_{\text{regular}} = \lambda_{\text{lbs}} \mathcal{L}_{\text{lbs}} + \lambda_{\text{smpl}} \mathcal{L}_{\text{smpl}} + \lambda_{\text{dyn}} \mathcal{L}_{\text{dyn}}
\end{equation}
The term $\mathcal{L}_{\text{smpl}}$ is the $\mathcal{L}_1$ loss measuring the deviation of foreground Gaussian points from the reconstructed SMPL surface. $\mathcal{L}_{\text{dyn}}$ is the $\mathcal{L}_2$ loss applied to all dynamic attributes predicted by the deformation network. We use the following weights: $\lambda_{\text{lbs}} = 20$, $\lambda_{\text{smpl}} = 0.005$, and $\lambda_{\text{dyn}} = 0.01$.

\input{tables/pre_scene_neuman}
\input{tables/supply_scene_reconstruction}
\input{figures/scene_qualitative}
\section{Supplementary Experiments}
\label{sec:Supplementary_Experiments}
\subsection{Per-Scene Experimental Results}
We provide a comprehensive scene-by-scene performance comparison on the NeuMan dataset, as detailed in Table~\ref{tab:pre_scene_neuman} and Table~\ref{tab:scene_reconstruction}. To offer a multi-faceted evaluation, we report metrics under two distinct settings. First, Table~\ref{tab:pre_scene_neuman} presents the results calculated exclusively within the \textbf{human-only regions}, which are cropped using tight bounding boxes for each sequence to directly reflect the fidelity of our human reconstruction. Second, Table~\ref{tab:scene_reconstruction} provides the evaluation on the \textbf{entire images} to assess the overall scene reconstruction quality, including the background. 

Both quantitative evaluations employ standard photometric metrics (PSNR, SSIM~\cite{wang2004image}, and LPIPS~\cite{zhang2018unreasonable}) as defined in \S\ref{sec:Dataset}. As shown in these results, our approach consistently outperforms existing SOTA methods across various complex scenes. Visual results in Figure~\ref{fig:scene_qualitative} further confirm that our model achieves superior spatial alignment and texture detail for both the foreground human and the static background.

\input{figures/smpl_vis}
\subsection{Visualization of Human Pose Optimization}
Beyond pixel-level photometric evaluations, we further assess the effectiveness of our joint optimization by visualizing the refinement of human pose parameters. In Figure~\ref{fig:smpl_vis}, we overlay the estimated 3D human meshes onto the ground truth images to qualitatively evaluate the spatial alignment. 

It is evident that our method successfully corrects significant misalignments present in the initial poses. Quantitatively, our joint optimization reduces the Mean Per-Joint Position Error (MPJPE) by 4mm compared to the initial estimates. As shown in the middle column of Figure~\ref{fig:smpl_vis}, our optimized results demonstrate precise alignment with the image evidence, whereas the unoptimized initial estimates (right column) exhibit noticeable deviations from the actual human contours. By iteratively refining the body pose $\{\boldsymbol{\theta}_t, \boldsymbol{\beta}\}$ in tandem with camera trajectories, our model achieves superior geometric precision. This accurately aligned pose provides a stable and consistent anchor for the Gaussian attributes, thereby ensuring the temporal coherence of the reconstructed human avatars.

\subsection{Demonstration Video}
We provide a demonstration video on our GitHub project page to visually showcase the high fidelity of our reconstruction. The video includes extensive results for both novel-view and novel-pose synthesis, confirming the effectiveness of our approach in modeling dynamic human bodies within the scene.

%% file: figures/ablation_stage.tex
\begin{figure*}[h]
    \centering
    \includegraphics[width=\linewidth]{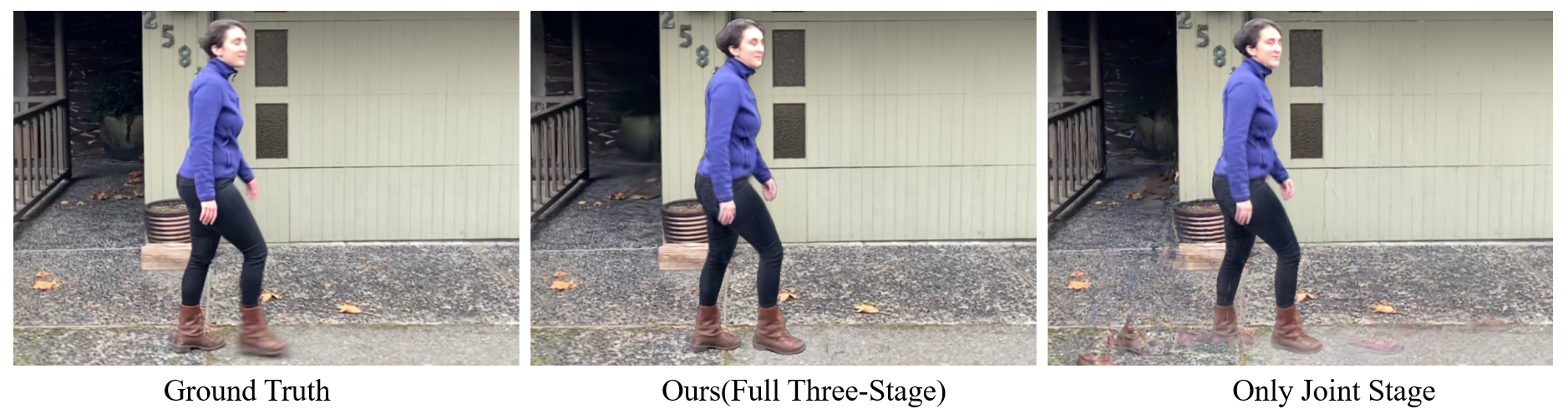}
    \caption{\textbf{Ablation study on the phased optimization schedule.} From left to right: the Ground Truth reference, results of our full three-stage optimization schedule, and results of performing joint optimization only (without warm-up and independent stages).}
    \label{fig:ablation_stage}
\end{figure*}

%% file: figures/model_architecture.tex
\begin{figure*}[t]
    \centering
    \includegraphics[width=\linewidth]{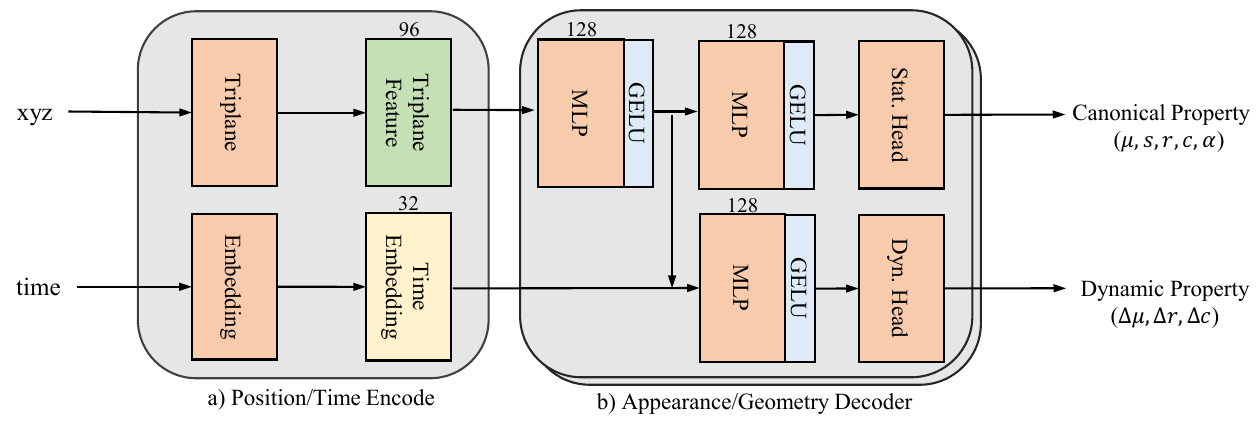}

    \caption{\textbf{JOintGS Model Architecture.} Our model architecture is composed of one Encoder module and two Decoder modules. In the Encoder module, the position attributes and the global temporal attributes of the Gaussian functions are encoded into positional features and temporal features, respectively. The Decoder module receives the positional features as input and utilizes a two-layer MLP with GELU activation to output either appearance or geometry features. These features are then fed into corresponding prediction heads to derive the specific attribute values. For certain dynamic attributes, we opt to inject the temporal features into the second layer of the MLP and use the same prediction head to output the corresponding residual values.
    }
    \label{fig:model_architecture}
\end{figure*}

%% file: tables/pre_scene_neuman.tex
\definecolor{tabfirst}{rgb}{1, 0.7, 0.7}
\definecolor{tabsecond}{rgb}{1, 0.85, 0.7}
\definecolor{tabthird}{rgb}{1, 1, 0.7}

\begin{table*}[htb!]
    \centering
    \resizebox{\textwidth}{!}{
    \begin{tabular}{c|ccc|ccc|ccc|ccc|ccc|ccc}
    \toprule
        & \multicolumn{3}{c|}{\textbf{Seattle}} & \multicolumn{3}{c|}{\textbf{Citron}} & \multicolumn{3}{c|}{\textbf{Parking}} & \multicolumn{3}{c|}{\textbf{Bike}} & \multicolumn{3}{c|}{\textbf{Jogging}} & \multicolumn{3}{c}{\textbf{Lab}}   \\
    \midrule
         & \footnotesize{PSNR $\uparrow$} & \footnotesize{SSIM $\uparrow$} & \footnotesize{LPIPS $\downarrow$} & \footnotesize{PSNR $\uparrow$} & \footnotesize{SSIM $\uparrow$} & \footnotesize{LPIPS $\downarrow$} & \footnotesize{PSNR $\uparrow$} & \footnotesize{SSIM $\uparrow$} & \footnotesize{LPIPS $\downarrow$} & \footnotesize{PSNR $\uparrow$} & \footnotesize{SSIM $\uparrow$} & \footnotesize{LPIPS $\downarrow$} & \footnotesize{PSNR $\uparrow$} & \footnotesize{SSIM $\uparrow$} & \footnotesize{LPIPS $\downarrow$} & \footnotesize{PSNR $\uparrow$} & \footnotesize{SSIM $\uparrow$} & \footnotesize{LPIPS $\downarrow$} \\
    \midrule 
    Vid2Avatar~\cite{guo2023vid2avatar} 
    &  16.90 &  0.51 &  0.27 
    &  15.96 &  0.59 &  0.28 
    & \cellcolor{tabthird}18.51 &  0.65 & 0.26 
    &  12.44 &  0.39 & 0.54 
    &  16.36 &  0.46 &  0.30 
    &  15.99 &  0.62 & 0.34 \\
    NeuMan~\cite{jiang2022neuman}     
    & \cellcolor{tabthird}18.42 & \cellcolor{tabthird}0.58 & \cellcolor{tabthird}0.20 
    & \cellcolor{tabthird}18.39 & \cellcolor{tabthird}0.64 & \cellcolor{tabthird}0.19 
    &  17.66 & \cellcolor{tabthird}0.66 & \cellcolor{tabthird}0.24 
    & \cellcolor{tabthird}19.05 & \cellcolor{tabthird}0.66 & \cellcolor{tabthird}0.21 
    &  \cellcolor{tabsecond}17.57 & \cellcolor{tabthird}0.54 & \cellcolor{tabthird}0.29 
    & \cellcolor{tabthird}18.76 & \cellcolor{tabthird}0.73 & \cellcolor{tabthird}0.23 \\
    HUGS~\cite{kocabas2024hugs}       
    &  \cellcolor{tabsecond}19.06 &  \cellcolor{tabsecond}0.67 &  \cellcolor{tabsecond}0.15 
    &  \cellcolor{tabsecond}19.16 &  \cellcolor{tabsecond}0.71 &  \cellcolor{tabsecond}0.16 
    &  \cellcolor{tabsecond}19.44 &  \cellcolor{tabsecond}0.73 &  \cellcolor{tabsecond}0.17 
    &  \cellcolor{tabsecond}19.48 &  \cellcolor{tabsecond}0.67 &  \cellcolor{tabsecond}0.18 
    & \cellcolor{tabthird}17.45 &  \cellcolor{tabsecond}0.59 &  \cellcolor{tabsecond}0.27 
    &  \cellcolor{tabsecond}18.79 &  \cellcolor{tabsecond}0.76 &  \cellcolor{tabsecond}0.18\\
    \midrule
    JOintGS (Ours)       
    &  \cellcolor{tabfirst}25.13 &  \cellcolor{tabfirst}0.88 &  \cellcolor{tabfirst}0.08 
    &  \cellcolor{tabfirst}25.39 &  \cellcolor{tabfirst}0.87 &  \cellcolor{tabfirst}0.10 
    &  \cellcolor{tabfirst}24.70 &  \cellcolor{tabfirst}0.86 &  \cellcolor{tabfirst}0.17 
    &  \cellcolor{tabfirst}25.21 &  \cellcolor{tabfirst}0.85 &  \cellcolor{tabfirst}0.16 
    & \cellcolor{tabfirst}21.77 &  \cellcolor{tabfirst}0.77 &  \cellcolor{tabfirst}0.16 
    &  \cellcolor{tabfirst}25.30 &  \cellcolor{tabfirst}0.87 &  \cellcolor{tabfirst}0.14\\
    \bottomrule
    \end{tabular}  
    }
    \caption{
    \textbf{Human reconstruction quality on NeuMan dataset.} Quantitative comparison on human-only regions cropped using tight bounding boxes. Our method achieves state-of-the-art performance across all six sequences
    }
    \label{tab:pre_scene_neuman}
\end{table*}

%% file: tables/supply_scene_reconstruction.tex
\definecolor{tabfirst}{rgb}{1, 0.7, 0.7}
\definecolor{tabsecond}{rgb}{1, 0.85, 0.7}
\definecolor{tabthird}{rgb}{1, 1, 0.7}

\begin{table*}[htb!]
    \centering
    \resizebox{\textwidth}{!}{
    \begin{tabular}{c|ccc|ccc|ccc|ccc|ccc|ccc}
    \toprule
        & \multicolumn{3}{c|}{\textbf{Seattle}} & \multicolumn{3}{c|}{\textbf{Citron}} & \multicolumn{3}{c|}{\textbf{Parking}} & \multicolumn{3}{c|}{\textbf{Bike}} & \multicolumn{3}{c|}{\textbf{Jogging}} & \multicolumn{3}{c}{\textbf{Lab}}   \\
    \midrule
         & \footnotesize{PSNR $\uparrow$} & \footnotesize{SSIM $\uparrow$} & \footnotesize{LPIPS $\downarrow$} & \footnotesize{PSNR $\uparrow$} & \footnotesize{SSIM $\uparrow$} & \footnotesize{LPIPS $\downarrow$} & \footnotesize{PSNR $\uparrow$} & \footnotesize{SSIM $\uparrow$} & \footnotesize{LPIPS $\downarrow$} & \footnotesize{PSNR $\uparrow$} & \footnotesize{SSIM $\uparrow$} & \footnotesize{LPIPS $\downarrow$} & \footnotesize{PSNR $\uparrow$} & \footnotesize{SSIM $\uparrow$} & \footnotesize{LPIPS $\downarrow$} & \footnotesize{PSNR $\uparrow$} & \footnotesize{SSIM $\uparrow$} & \footnotesize{LPIPS $\downarrow$} \\
    \midrule 

    HyperNeRF~\cite{park2021hypernerf}
    & 16.43 & 0.43 & 0.40
    & 16.81 & 0.41 & 0.56
    & 16.04 & 0.38 & 0.62
    & 17.64 & 0.42 & 0.43
    & 18.52 & 0.39 & 0.52
    & 16.75 & 0.51 & 0.23 \\
    Vid2Avatar~\cite{guo2023vid2avatar}
    & 17.41 & 0.56 & 0.60
    & 14.32 & 0.62 & 0.65
    & 21.56 & 0.69 & 0.50
    & 14.86 & 0.51 & 0.69
    & 15.04 & 0.41 & 0.70
    & 13.96 & 0.60 & 0.68 \\
    NeuMan~\cite{jiang2022neuman}
    & \cellcolor{tabthird}23.99 & \cellcolor{tabthird}0.78 & \cellcolor{tabthird}0.26
    & \cellcolor{tabthird}24.63 & \cellcolor{tabthird}0.81 & \cellcolor{tabthird}0.26
    & \cellcolor{tabthird}25.43 & \cellcolor{tabthird}0.80 & \cellcolor{tabthird}0.31
    & \cellcolor{tabsecond}25.55 & \cellcolor{tabthird}0.83 & \cellcolor{tabthird}0.23
    & \cellcolor{tabthird}22.70 & \cellcolor{tabthird}0.68 & \cellcolor{tabthird}0.32
    & \cellcolor{tabthird}24.96 & \cellcolor{tabthird}0.86 & \cellcolor{tabthird}0.21 \\

    HUGS~\cite{kocabas2024hugs}
    & \cellcolor{tabsecond}25.94 & \cellcolor{tabsecond}0.85 & \cellcolor{tabsecond}0.13
    & \cellcolor{tabsecond}25.54 & \cellcolor{tabsecond}0.86 & \cellcolor{tabsecond}0.15
    & \cellcolor{tabsecond}26.86 & \cellcolor{tabsecond}0.85 & \cellcolor{tabsecond}0.22
    & \cellcolor{tabthird}25.46 & \cellcolor{tabsecond}0.84 & \cellcolor{tabsecond}0.13
    & \cellcolor{tabsecond}23.75 & \cellcolor{tabsecond}0.78 & \cellcolor{tabsecond}0.22
    & \cellcolor{tabsecond}26.00 & \cellcolor{tabsecond}0.92 & \cellcolor{tabsecond}0.09 \\
    
    \midrule
    JOintGS (Ours)
    & \cellcolor{tabfirst}32.52 & \cellcolor{tabfirst}0.95 & \cellcolor{tabfirst}0.04
    & \cellcolor{tabfirst}26.27 & \cellcolor{tabfirst}0.84 & \cellcolor{tabfirst}0.09
    & \cellcolor{tabfirst}32.77 & \cellcolor{tabfirst}0.92 & \cellcolor{tabfirst}0.09
    & \cellcolor{tabfirst}31.19 & \cellcolor{tabfirst}0.94 & \cellcolor{tabfirst}0.04
    & \cellcolor{tabfirst}28.29 & \cellcolor{tabfirst}0.89 & \cellcolor{tabfirst}0.11
    & \cellcolor{tabfirst}30.35 & \cellcolor{tabfirst}0.94 & \cellcolor{tabfirst}0.06\\
    \bottomrule
    \end{tabular}  
    }
    \caption{
    \textbf{Full-image reconstruction quality assessment on the NeuMan dataset.} Evaluation is conducted on the entire frames without any region-specific cropping or masking.
    }
    \label{tab:scene_reconstruction}
\end{table*}

%% file: figures/scene_qualitative.tex
\begin{figure*}[t]
    \centering
    \includegraphics[width=\linewidth]{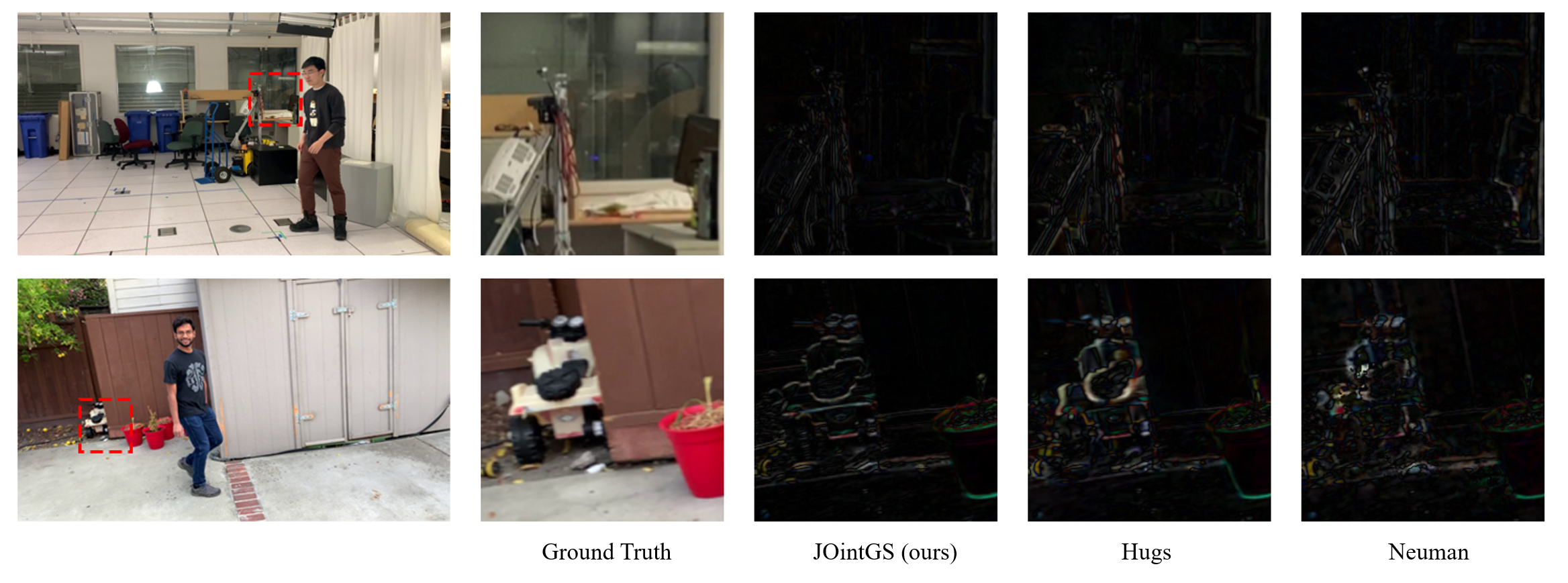}
    \caption{\textbf{Qualitative comparison on NeuMan dataset.} For each scene, we present the complete rendered image (first column) and a zoomed-in view of a densely textured region (second column), along with the error map of different methods (last three columns).} 
    \label{fig:scene_qualitative}
\end{figure*}

%% file: figures/smpl_vis.tex
\begin{figure*}[t]
    \centering
    \includegraphics[width=\linewidth]{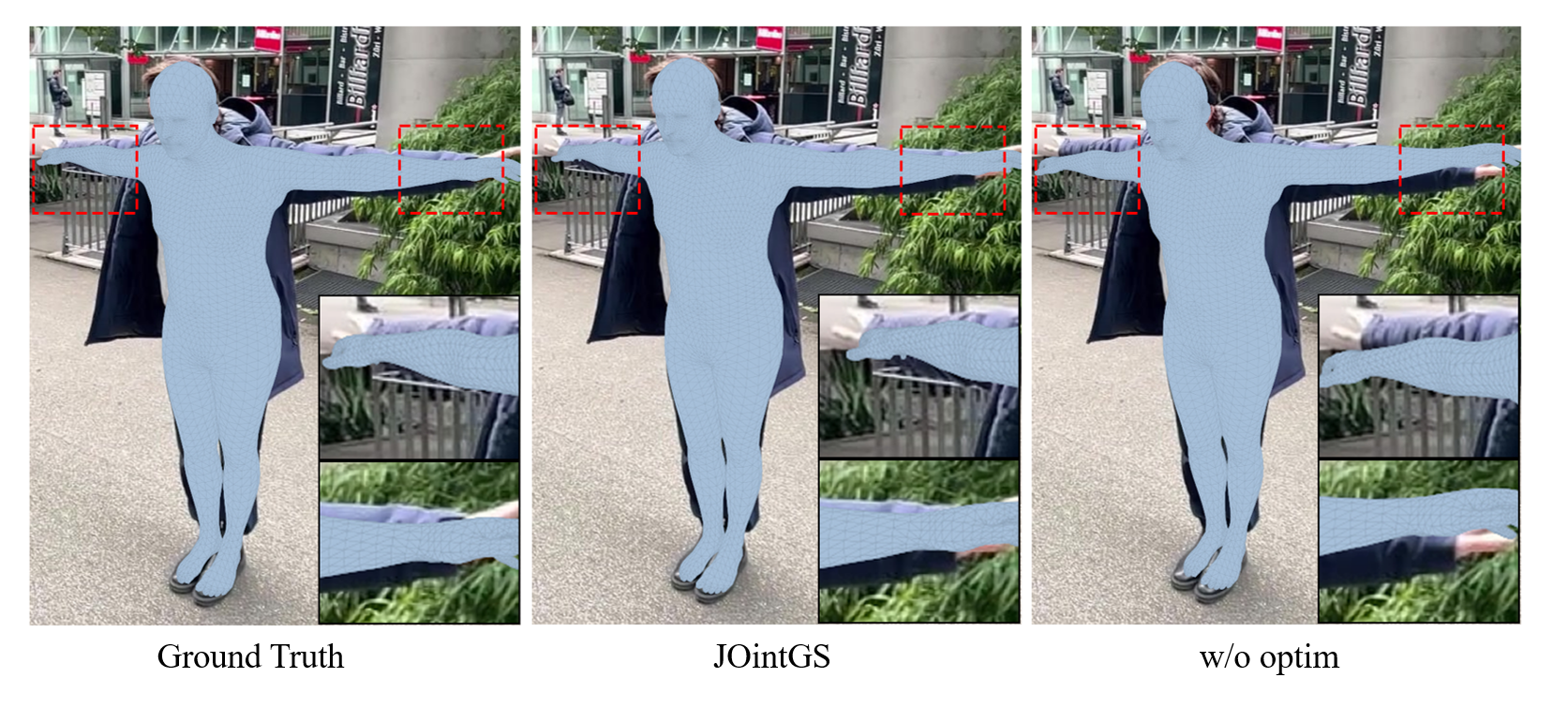}
    \caption{\textbf{Qualitative comparison of SMPL pose refinement.} We overlay the estimated 3D human meshes onto the ground truth images to evaluate spatial alignment. From left to right: (Left) The reference image with ground truth pose; (Middle) Results after our JOintGS optimization; (Right) Results without optimization (initial estimates).}
    \label{fig:smpl_vis}
\end{figure*}